# Distant Pedestrian Detection in the Wild using Single Shot Detector with Deep Convolutional Generative Adversarial Networks

Ranjith Dinakaran[1], Philip Easom[1], Li Zhang[1], Ahmed Bouridane[1], Richard Jiang[2], Eran Edirisinghe[3]

[1]*Computer & Information Sciences, Northumbria University, Newcastle upon Tyne, UK*
[2]*Computing & Communication, Lancaster University, Lancaster, UK*
[3]*Computer Science, Loughborough University, Loughborough, UK*

*Abstract*— **In this work, we examine the feasibility of applying Deep Convolutional Generative Adversarial Networks (DCGANs) with Single Shot Detector (SSD) as data-processing technique to handle with the challenge of pedestrian detection in the wild. Specifically, we attempted to use in-fill completion (where a portion of the image is masked) to generate random transformations of images with portions missing to expand existing labelled datasets. In our work, GAN's been trained intensively on low resolution images, in order to neutralize the challenges of the pedestrian detection in the wild, and considered humans, and few other classes for detection in smart cities. The object detector experiment performed by training GAN model along with SSD provided a substantial improvement in the results. This approach presents a very interesting overview in the current state of art on GAN networks for object detection. We used Canadian Institute for Advanced Research (CIFAR), Caltech, KITTI data set for training and testing the network under different resolutions and the experimental results with comparison been showed between DCGAN cascaded with SSD and SSD itself.**

## I. INTRODUCTION

Generative Adversarial Networks (GANs) [1] have been of specific interest in the deep learning paradigm, especially for image processing, synthesis and generation. A high-level formulation of this task is to learn how to create realistic data based on ground truth. The problem is disintegrated into two networks. A generator network is in charge for generating an image given from a randomly generated source of noises (e.g. a noise vector). A discriminator is in charge to identify apart a real image (from a large corpus), and the randomly generated example from the generator. In simplest way the problem we dealt with is, a gradient is computed based on the classification and, the loss is backpropagated through generator and discriminator networks. Our work investigates the feasibility of applying GANs to datasets in object detection for smart cities. We provide GAN a set of visual objects from CIFAR dataset known to appear in an image and providing the task to the GAN with arranging the visual object and filling in the context around them in a realistic manner. The discriminator is assigned with the task of discriminating between the real image containing the visual objects provided in the dataset, and the generated image, with the generated background pixels between the visual objects.

The main motivation of this work starts from the training data for object detection [1, 18-23] is often it is harder to collect the data and it is much harder to compare and classify the data, (using variations of existing labelled data to artificially increase dataset size) also it is a common technique in supervised learning. In our work, we mainly focus on GANs especially in wide amount of data technique to improve an existing object detection task.

## II. RELATED WORK

### A. Generative Adversarial Networks

The Generative Adversarial Networks (GANs) [1] is an outline for learning generative models. Mathieu et al. [9] and Dentonet al. [10] adopted GANs for the application of image generation. In [11] and [12], GANs were employed to learn a mapping from one manifold to another for style transfer and inpainting, respectively. The idea of using GANs for unsupervised representation learning was described in [13]. GANs were also applied to image super resolution in [14]. To the best of our knowledge, this work makes the first attempt to accommodate GANs on the object detection task to address the small-scale problem by generating super-resolved representations for small objects.

### B. Generative Models

Recently several attempts have been made to improve image generation using generative models. The most popular generative model approaches are Generative Adversarial Networks (GANs) [1], Variational Autoencoders (VAEs) [6], and Autoregressive models [7]. And their variants, e.g. conditional GANs, reciprocative Conditional GANs, Deep Convolutional GANs (DC-GANs) [8], etc. Radford et al. [8] use a Conv-Deconv GAN architecture to learn good image representation for several image synthesis tasks. Denton et al. [9] use a Laplacian pyramid of generators and discriminators to synthesize multi-scale high resolution images. Mirza and Osindero [10] train GANs by explicitly providing a conditional variable to both the generator and the discriminator, using one-hot encoding to control generated image features, namely conditional GANs (cGANs). Reed et al. [11] use a DC-GAN conditioned on text features encoded by a hybrid character-level convolutional RNN. Perarnau et al. [2] use an encoder with a conditional GAN (cGAN), to inverse the mapping of a cGAN for complex image editing, calling the result Invertible cGANs. Dumoulin et al. [12] and Donahue et al. [13] use an encoder with GANs. Makhzani et al [14] and Larsen et al. [15] use a similar idea to [2] but combining a VAE and GAN to improve the realism of the generated images.

## III. PRELIMINARIES ON DCGAN

### A. Generative Adversarial Networks

Let y be the original input image, and y' represent the generated image from the generative model give a noise vector sampled from x ~ uniform (0.1). Let D(z) and D(z') be the output of the discriminator network for the two images and represent the regressed confidence that the input is from the number of real images. Intuitively, we seek to minimize D(z') and maximize D(z). The value function id D that we can write is now:

$$\log(D(x)) + \log(1-D(z')) \quad (1)$$

The complete minimax value function V for G and D now becomes as defined by Goodfellow [1]:

$$\min \max V(D, G) = E_z(\log D(z)) + E_{z'}(\log(1-D(z'))) \quad (2)$$

The challenging task is image decryption. Whereas G can choose any point in the generator multiple images to consider a valid image, in the decryption task, we first consider G with a set of predetermined pixels based on ground truth, these pixels can be a random pixel. The experiment becomes very challenging because now a very small number of multiple images will produce a accurate image given to the existing images. This is typically achieving by backpropagation through a multi-component reconstruction loss.

Fig 1 shows the architecture used for the generative portion of this system. The noise vector x is expanded through a series of deconvolutional layers. Our Network input is 19x19 dimension where the labelled visual objects are retained in original configuration, and the left-over pixels are sampled and resampled from noise distribution heuristically Gaussian approach. The image experiences a series of convolutional layers while its retaining the same dimensions., we found that in this resolution GANs generator part will not congregate nor begin to learn which is appropriate in data setting.

### B. DCGAN based Distant Pedestrian Rendering

Fig.2 shows two columns with images of low resolution raw images, and high resolution images which is processed and enhanced by GAN. The raw images obtained from the video, with a low pixel ratio of 320X100, and with the size of the 27kb, on the other hand we had a high resolution images where the pixels and the ratio of the image been enhanced, after enhancement 1920X1080, after enhancement the size of the image also been enhanced to 451kb, now the enhanced images will present very good detection which is shown in the Fig 6, also the Fig 5 shows the detections in the images taken from the different datasets, the main advantage in this process is the processing time in detection is reduced, maximum output in detection with the help of enhancement in images is done by filling the latent space using GAN.

The input for generator is given a bit of higher features from the images extracted from dataset, which is also an advantage for the exact image generation which we used from dataset, we can clearly see the images in low resolution from Fig 2, where only the features only been fed

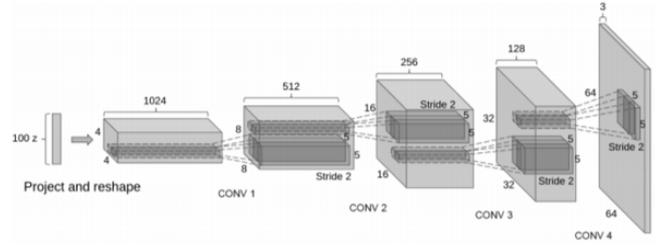

Fig 1. The DCGAN architecture.

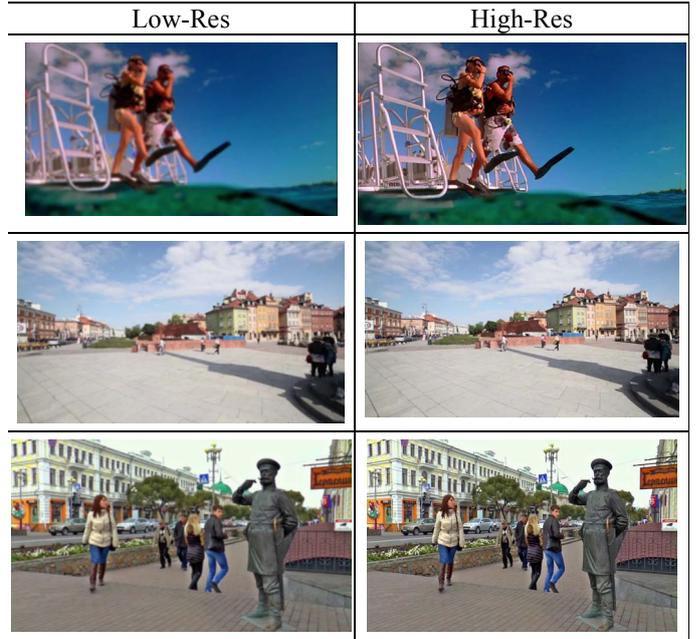

| Low-Res | High-Res |

Fig2. The generated high-res images from low-res ones.

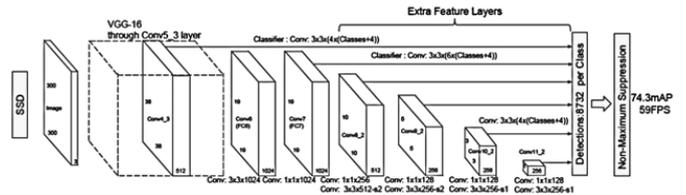

Fig3. The SSD architecture.

to the generator and the same image is given to the adversarial part for discriminator, on the first hand the output of the GAN is very efficient with 100%.

## IV. PROPOSED DCGAN-SSD ARCHITECTURE FOR PEDESTRIAN DETECTION

### A. Single Shot Detector

The SSD method we used in is based on a feed-forward convolutional network that produces a fixed-size collection of bounding boxes and scores for the presence of object class instances in those boxes, followed by a non-maximum suppression step to produce the final detections. The early network layers are based on a standard architecture used for high quality image classification (truncated before any classification layers), which we will call the base network2.

We then add secondary structure to the network to produce detections with the following key features:

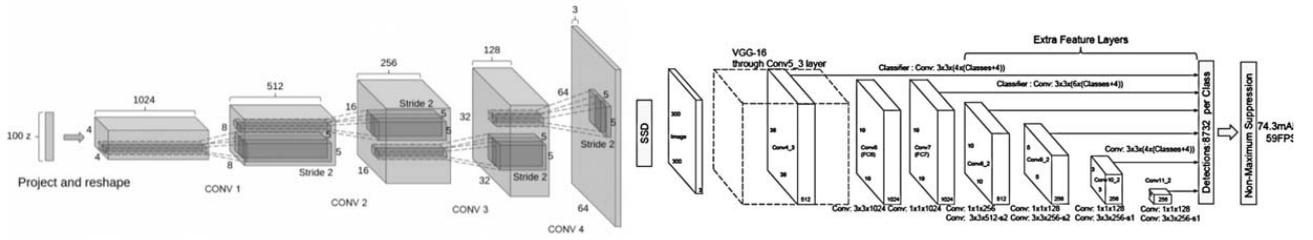

Fig 4. The Architecture of DCGAN + SSD.

In the convolution layers of SSD, each added feature layer (or optionally an existing feature layer from the base network) can produce a fixed set of detection predictions using a set of convolutional filters. These are indicated on top of the SSD network architecture in Fig. 4. For a feature layer of size m × n with p channels, the basic element for predicting parameters of a potential detection is a 3 × 3 × p small kernel that produces either a score for a category, or a shape offset relative to the default box coordinates. At each of the m × n locations where the kernel is applied, it produces an output value. The bounding box offset output values are measured relative to a default box position relative to each feature map location.

*B. DCGAN Enhanced SSD*

Object detection is considered as a major challenge of the image classification task, where the main goal is to classify and localize every object from input image. The object detection problem is considered as a major challenge in computer vision and, made some progress in recent years because of advanced machine learning tools like deep learning and GANs [1]. The main advances of region proposal methods are two state of art methods You only look once (YOLO) and SSD [4], improved the object detection in computer vision. When we cascade DCGAN+SSD the main advantage is, the combination reduces the scale factors in the images, and also with SSD we add convolutional feature layers to the end of curtailed base network. These layers in GAN and SSD reduce the size progressively at different scales. The model of making predictions varies for each feature layer that operates in SSD.

The most communal technique for validating the eminence of unsupervised illustration learning algorithms is to apply them as a feature extractor on supervised datasets and validate the performance of linear models tailored on top of these features as shown in Fig 3 the convolutional layers are build to resize the images and there is no room for enhancement of images or to fill the feature space. However the Fig 4 demonstrates the changes in the convolutional layers and these layers in the GAN is build to enhance the image quality under any size, so that the feature space is filled with feature maps, this help in better detection quality.

With datasets we used gave a strong starting point performance has been verified from a well-tuned single layer feature extraction pipeline[1]. When using a high amount of feature maps (4800), to compromise the latent space with the help of GAN and the detection through SSD, this technique achieves 80.7% of accuracy. An unsupervised multi-layered and multi box detection model using DCGAN & SSD with base algorithm gave us nearly 90% accuracy. To validate the quality of the depictions learned by DCGANs for supervised tasks, we use pretrained SSD which has been trained with ImageNet dataset, so SSD is basically a pretrained network what we used in our experiment. We use the pretrained network to save the computational time, and then use the discriminator's convolutional features from all layers, maxpooling each layers representation to produce a 4 × 4 spatial grid for 32x32 size images. These features are then flattened and concatenated to form a dimensional vector and a regularized linear classifier is trained on top of them. This achieves better accuracy, than by using SSD on its own, we like to make a point even the SSD is a very good detector, but when it combines with other classifiers, it performs much better. Notably, the SSD detector performance lags in the scale factor, which is compromised by DCGAN in ou experiment, where the DCGAN can improvise the scale factor in SSD by providing the detector with super resolution images, it does result in a larger total feature vector size due to the highest layers given for feature vectors of 4 × 4 spatial locations. The performance of DCGANs is still less than that of Exemplar CNNs [16], a technique which trains normal discriminative CNNs in an unsupervised fashion to differentiate between specifically chosen, aggressively augmented, exemplar samples from the source dataset. Further improvements could be made by finetuning the discriminator's representations, but we leave this for future work. Additionally, since our DCGAN was never trained on CIFAR-10 this experiment also demonstrates the domain robustness of the learned features.

In Convolutional detector, each feature layer can absolutely produce a fixed set of predictions in detection with the help of convolutional filters. These filters have been mentioned on top of the SSD architecture in Fig.1, SSD were associated with a set of bounding boxes which it got default sizes with each cell from each feature map. The default size bounding boxes tile in a convolutional manner with the feature map, so that the position of each bounding box is related to its sizes in the cell, and also the class indicate the presence of feature cell in the bounding boxes.

*C. More Details of the Proposed DCGAN +SSD*

Leaky ReLU (negative_slope = 0.01, inplace = false) it applies element-wise

$$f(x) = \max(0, x) + negative_{slope} \times \min(0, x) \quad (3)$$

negative_slope – controls the angle of the negative slope by default 1e-2 inplace – can optionally do the operation inplace by default false.

Basically for the convolutions the Leaky ReLU works better than simple ReLU, we use Leaky ReLU for our consecutive generator layers and we use sigmoid rectification for the output of the discriminator. A discriminator is the second brain of our network and, the generator is the first brain of our network. We set a new target to 1 always and, calculate the loss between the output of the discriminator value between 0 and 1, the error is back propagated in generator not in discriminator, to update the weight of generator neural network.

To measure the error value between 0 and 1 we use a different criterion, based on ground truth that will be only between 0 and 1.

Therefore, the criterion of neural network is

Criterion = nn.CE Loss

The CE Loss is nothing but, it creates criterion that measures the Cross Entropy between the target and the output.

$$Loss(o,t) = -\frac{1}{n}\sum_{i}(t[i]) + (1 - t[i]) * \log(1 - o[i])) \quad (4)$$

Or in the case of the weights arrangements being specified, or in case of pre-trained network.

$$Loss(o,t) = -\frac{1}{n}\sum_{i} weights[i] * (t[i] * (1 - t[i]) * \log(1 - o[i])) \quad (5)$$

This is used for measuring the error of reconstruction for example an auto encoder which we used in our experiments, Note: the target t[i] should be numbers between 0 and 1.

The Experiment undergone 25 epochs for our images and SSD is given to Epoch 25 for object detection in humans, vehicles and animals.

## V. EXPERIMENTRAL RESULTS

*A. Experimental Condition*

This experiment uses Canadian Institute For Advanced Research 10 (CIFAR 10) dataset. The validation task for this dataset is to analyse the detection quality through bounding box evaluation image dataset. CIFAR 10 dataset contains 60000 objects from 89 categories, this dataset approximately consists of 4000 detected visual objects, across approximately 2000 images. The specific metric will optimize for the F1 score used as a detection threshold of 0.5, to avoid occlusion problem, and non – maximum suppression threshold of 0.2. The non-maximum suppression is important in detection where there are multiple bounding boxes detects the same objects multiple times. A very common approach to improve this is to sort out the bounding boxes by its score and squeeze the bounding boxes with its least score and an overlap of least non-maximum suppression threshold, which is 0.2 for our experiment in cooperation with our existing result. In accordance with the F1 score, accuracy has been reported for our baseline model in Table 1. We also shown that the larger datasets can also be improved significantly with our score quality.

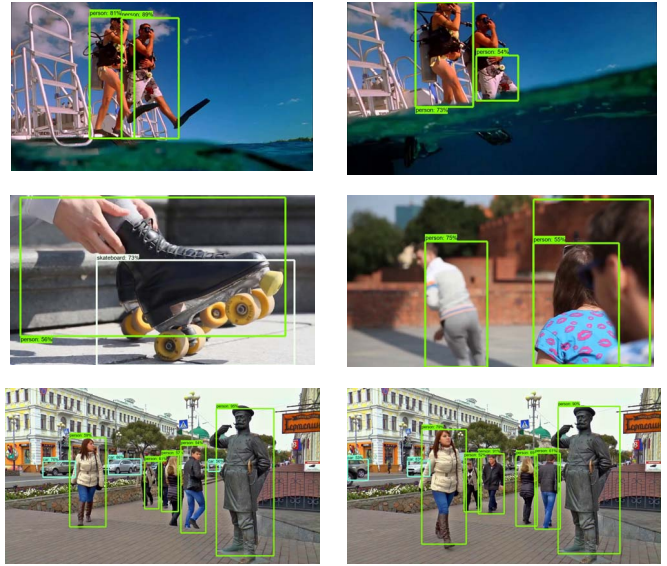

Fig.5 Sample images from different dataset's with annotations.

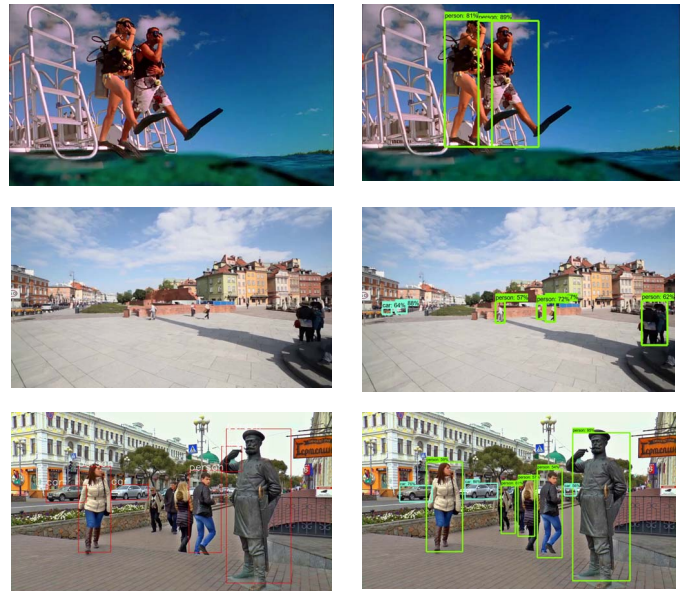

(a) SSD only      (b) DCGAN+SSD

Fig.6 Detection examples for comparison. (a) Results with SSD only; b) Results with DCGAN+SSD.

Table 1. Experimental results for comparison with datasets

| Method | CALTECH | CIFAR | KITTI | VOC |
|---|---|---|---|---|
| DCGAN+SSD | 93.6 | 87.9 | 92.3 | 88.9 |
| SSD | 65.8 | 53.5 | 58.3 | 49.5 |

The experimental results the working of DCGAN under low resolution images, where the DCGAN upgrades the pixels and brings it up to the detection quality before the images are fed to SSD for better detection. As a part of the

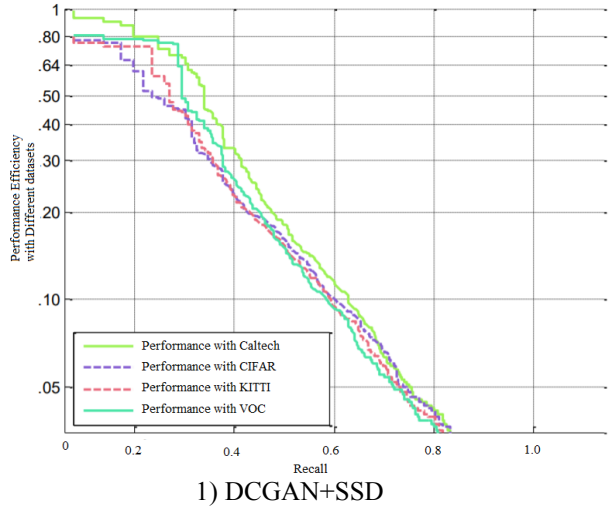

1) DCGAN+SSD

2) SSD only

Fig.7 Test results on precision and recall rates on different datasets.

experiment, we identified the difference when SSD works on its own and when it works with the DCGAN, the Fig 6. Shows the difference between the detection only with SSD and with DCGAN + SSD.

The Generative Adversarial Network was implemented on two different tasks: first we implemented to recreate the state of art GAN results using a multiple object encoding, to identify the objects and recreate the object using the object encoder by using GAN. We trained a GAN on CIFAR, Caltech, KITTI, VOC dataset, Fig.7 shows the performance of DCGAN+SSD and SSD only with different datasets, using an implementation with image sizes of 32x32, along with the batch size of 72, and 25 epochs, across a total of 60,000 images. We then applied the encoding task as described in the previous section, with the results in Fig 6, where we demonstrate the high reliability photo-realism between 0 to 25 of GAN training, specifically, we implemented Pytorch and Tensorflow implementation of deep convolutional GAN (DCGAN) as mentioned in [6].

During the experiment the research underwent many issues on attempting to train a GAN model on images under different resolutions larger than 64x64 and on resolutions like 100x100, and 128x128, there is no convergence in the generator model. Eventually we identify that it was not feasible to achieve the original data processing goal.

*B. Details of Datasets*

The CIFAR, Caltech, KITTI dataset contains of 60000 32x32 colour images in 10 classes, with more than 6000 images per class. It consist of 50000 training images and 10000 test images, Each file contains 10000 such 3073-byte "rows" of images, although there is nothing delimiting the rows. Therefore each file should be exactly 30730000 bytes long. [17].

The dataset is alienated into five training sets and one test set, each with 10000 images. The test set encompasses 1000 random images from each class. The training sets holds the remaining images in different order, but some training sets may consists of more images from one class when compare to another class. Between them, the training sets contain around 5000 images from each class. Fig.5 shows the sample images in the dataset.

*C. Experimental Results*

In our experiments, we used CIFAR, Caltech, and KITTI dataset to carry out our validation and examine if the proposed *GAN + SSD* architecture outperforms the single *SSD*, particularly on distant object or pedestrian detection.

In our experiments, DCGAN was implemented in two steps. First, we implemented the DCGAN codes based on PyTorch and Tensorflow, as illustrated in [6], which recreates the state-of-the-art GAN results using multiple object encoding. We trained our DCGAN on the CIFAR, Caltech, KITTI dataset, with all image size of 32×32, along with the batch size of 72, and 25 epochs, across a total of 60,000 images.

Fig.6 shows the detection results on the CIFAR, Caltech, KITTI, VOC dataset, using DCGAN+SSD and SSD only, respectively. Fig.6-a) is the results from SSD only, where a number of objects were missed in the detection. Fig.6-b) shows the results from DCGAN+SSD, where a number of missed objects in Fig.6-a) were detected successfully. The Fig 7 shows the recall ratio curve when we use the SSD only and DCGAN with SSD, the performance is notable. Particularly, objects at distance were detected in these images, which were mostly missed by the SSD only method.

In our experiments, DCGAN was implemented in two steps. First, we implemented the DCGAN codes based on PyTorch and Tensorflow, as illustrated in [6], which recreates the state-of-the-art GAN results using multiple object encoding. We trained our DCGAN on the CIFAR-10 ReLU for generator input and other respective layers of generator and for output from the generator we used Tanh in generator. For discriminator we use convolution itself, other parameters used in channels of the generator created by generated images, number of feature maps, kernel size, stride, padding, bias and for discriminator rectification we use Leaky ReLU, Leaky ReLu is very close to ReLU, but besides having max(0,x), it has negative slope multiplied by min(0,x).

## VI. CONCLUSION

In this work, we proposed a new architecture by cascading DCGANs with SSD to detect pedestrians and objects at distance, particularly for smart cities applications. With generative adversarial networks by implementing the DCGAN [6], more robust discriminative features are extracted around tiny objects and hence as a consequence, the detection rate is improved drastically. With such an apparent evidence to demonstrate its advantages, we can expect the proposed architecture will be valuable for smart cities applications that need to detect objects in the wild, particularly those tiny objects at distance, to secure life and avoid accidents. While GANs have been very successful in many other applications, our work successfully append it to a robust solution to the real world challenges in smart cities. The resulting models will also be deployed to other computer vision and transfer learning tasks such as image description generation [24], semantic image segmentation from dermoscopic and microscopic images [25, 26], and facial and gesture expression recognition in the wild [27].

## REFERENCES

[1] I. Goodfellow, J. Pouget-Abadie, M. Mirza, B. Xu, D. WardeFarley, S. Ozair, A. Courville, and Y. Bengio. Generative adversarial nets. In Advances in Neural Information Processing Systems, pages 2672–2680, 2014. 1, 2,3

[2] K. He, X. Zhang, S. Ren, and J. Sun. Deep residual learning for image recognition. arXiv preprint arXiv:1512.03385, 2015. 2

[3] G. B. Huang, M. Mattar, H. Lee, and E. Learned-Miller. Learning to align from scratch. In NIPS, 2012. 3

[4] W. Liu, D. Anguelov, D. Erhan, C. Szegedy, S. Reed, C.-Y. Fu, and A. C. Berg. Ssd: Single shot multibox detector. 2016. To appear. 2

[5] A. Nguyen, J. Yosinski, Y. Bengio, A. Dosovitskiy, and J. Clune. Plug & play generative networks: Conditional iterative generation of images in latent space. arXiv preprint arXiv:1612.00005, 2016. 4

[6] A. Radford, L. Metz, and S. Chintala. Unsupervised representation learning with deep onvolutional generative adversarial networks. arXiv preprint arXiv:1511.06434, 2015. 3, 4

[7] S. Ren, K. He, R. Girshick, and J. Sun. Faster R-CNN: Towards real-time object detection with region proposal networks. In Neural Information Processing Systems (NIPS), 2015. 2

[8] K. Simonyan and A. Zisserman. Very deep convolutional networks for large-scale image recognition. CoRR, abs/1409.1556, 2014. 2

[9] M. Mathieu, C. Couprie, and Y. LeCun. Deep multi-scale video prediction beyond mean square error. arXiv preprint arXiv:1511.05440, 2015. 2

[10] E. L. Denton, S. Chintala, R. Fergus, et al. Deep generative image models using a laplacian pyramid of adversarial networks. In NIPS, pages 1486–1494, 2015. 2

[11] C. Li and M. Wand. Combining markov random fields and convolutional neural networks for image synthesis. arXiv preprint arXiv:1601.04589, 2016. 2

[12] R. Yeh, C. Chen, T. Y. Lim, M. Hasegawa-Johnson, and M. N. Do. Semantic image inpainting with perceptual and contextual losses. arXiv preprint arXiv:1607.07539, 2016. 2

[13] A. Radford, L. Metz, and S. Chintala. Unsupervised representation learning with deep convolutional generative adversarial networks. arXiv preprint arXiv:1511.06434, 2015. 2

[14] C. Ledig, L. Theis, F. Huszar, J. Caballero, A. Aitken, A. Tejani, ́ J. Totz, Z. Wang, and W. Shi. Photo-realistic single image superresolution using a generative adversarial network. arXiv preprint arXiv:1609.04802, 2016. 2

[15] Larsen, A.B.L., Sønderby, S.K., Larochelle, H., Winther, O.: Autoencoding beyond pixels using a learned similarity metric. arXiv:1512.09300 (2015)[16] Dosovitskiy, Alexey, Fischer, Philipp, Springenberg, Jost Tobias, Riedmiller, Martin, and Brox, Thomas. Discriminative unsupervised feature learning with exemplar convolutional neural networks. In Pattern Analysis and Machine Intelligence, IEEE Transactions on, volume 99. IEEE, 2015.

[16] Dosovitskiy, Alexey, Fischer, Philipp, Springenberg, Jost Tobias, Riedmiller, Martin, and Brox, Thomas. Discriminative unsupervised feature learning with exemplar convolutional neural networks. In Pattern Analysis and Machine Intelligence, IEEE Transactions on, volume 99. IEEE, 2015.

[17] CIFAR -10 Available online :https://cs.toronto.edu/~kriz/cifar.html

[18] G. Storey, R. Jiang, A. Bouridane, "Role for 2D image generated 3D face models in the rehabilitation of facial palsy", IET Healthcare Technology Letters, 2017.

[19] G Storey, A Bouridane, R Jiang, "Integrated Deep Model for Face Detection and Landmark Localisation from 'in the wild' Images", IEEE Access, in press.

[20] R Jiang, ATS Ho, I Cheheb, N Al-Maadeed, S Al-Maadeed, A Bouridane, "Emotion recognition from scrambled facial images via many graph embedding", Pattern Recognition, Vol.67, July 2017.

[21] R Jiang, S Al-Maadeed, A Bouridane, D Crookes, M E Celebi, "Face recognition in the scrambled domain via salience-aware ensembles of many kernels", IEEE Trans. Information Forensics & Security, Vol.11, Aug 2016.

[22] R Jiang, A Bouridane, D Crookes, M E Celebi, H L Wei, "Privacy-protected facial biometric verification via fuzzy forest learning", IEEE Trans. Fuzzy Systems, Vol.24, Aug 2016.

[23] Ranjith Dinakaran, Richard Jiang, Graham Sexton, "Image Resolution Impact Analysis on Pedestrian Detection in Smart Cities Surveillance", ACM Proceedings. In IML, 2017.

[24] P. Kinghorn, L. Zhang, L. Shao, "A region-based image caption generator with refined descriptions". Neurocomputing. 272 (2018) 416-424.

[25] T. Tan, L. Zhang, S.C. Neoh, C.P. Lim, "Intelligent Skin Cancer Detection Using Enhanced Particle Swarm Optimization". Knowledge-Based Systems. 2018.

[26] W. Srisukkham, L. Zhang, S.C. Neoh, S. Todryk, C.P. Lim, "Intelligent Leukaemia Diagnosis with Bare-Bones PSO based Feature Optimization". Applied Soft Computing, 56. pp. 405-419. 2017.

[27] K. Mistry, L. Zhang, S.C. Neoh, C.P. Lim, B. Fielding, "A micro-GA Embedded PSO Feature Selection Approach to Intelligent Facial Emotion Recognition". IEEE Transactions on Cybernetics. 47 (6) 1496–1509. 2017.